\documentclass{article}

\usepackage{arxiv}

\usepackage[utf8]{inputenc}
\usepackage[T1]{fontenc}
\usepackage{microtype}
\usepackage{amsfonts, amsmath, amssymb, amsthm}
\usepackage{bm, bbm}
\usepackage{nicefrac}
\usepackage{booktabs}
\usepackage{graphicx}
\usepackage[font=small, belowskip=-1ex]{caption}
\usepackage[dvipsnames]{xcolor}
\usepackage[colorlinks=true, allcolors=blue]{hyperref}
\usepackage[authoryear, round, semicolon]{natbib}

\makeatletter
\def\blfootnote{\xdef\@thefnmark{}\@footnotetext}
\makeatother

\title{An Empirical Categorization of Prompting Techniques for Large Language Models: A Practitioner's Guide}
\author{
    Oluwole Fagbohun\\
    Tech Team\\
    Changeblock\\
    London, SW1H 0HW, UK\\
    \texttt{oluwole.fagbohun@changeblock.com}
\And
    Rachel M. Harrison\\
    GenAI Lab\\
    Ophiuchus LLC\\
    Dover, DE 19904, USA\\
    \texttt{rae@ophiuchus.ai}
\And
    Anton Dereventsov\\
    GenAI Lab\\
    Ophiuchus LLC\\
    Dover, DE 19904, USA\\
    \texttt{anton@ophiuchus.ai}
}

\renewcommand{\undertitle}{}
\renewcommand{\shorttitle}{An Empirical Categorization of Prompting Techniques for Large Language Models}

\hypersetup{
pdftitle={A Categorization of Prompting Techniques and Approaches for Large Language Models},
pdfsubject={cs.AI, cs.CL, cs.LG},
pdfauthor={Oluwole Fagbohun, Rachel Harrison, Anton Dereventsov},
pdfkeywords={large language models, LLMs, prompt engineering, prompt design, prompting techniques, ChatGPT},
}
\date{}

\begin{document}
\maketitle
\blfootnote{Preprint. This work was presented at the \href{https://unitedresearchforum.com/ai-conference/}{4th International Conference on AI, ML, Data Science, and Robotics (2023)}.}

\begin{abstract}
Due to rapid advancements in the development of Large Language Models (LLMs), programming these models with prompts has recently gained significant attention. However, the sheer number of available prompt engineering techniques creates an overwhelming landscape for practitioners looking to utilize these tools. For the most efficient and effective use of LLMs, it is important to compile a comprehensive list of prompting techniques and establish a standardized, interdisciplinary categorization framework. In this survey, we examine some of the most well-known prompting techniques from both academic and practical viewpoints and classify them into seven distinct categories. We present an overview of each category, aiming to clarify their unique contributions and showcase their practical applications in real-world examples in order to equip fellow practitioners with a structured framework for understanding and categorizing prompting techniques tailored to their specific domains. We believe that this approach will help simplify the complex landscape of prompt engineering and enable more effective utilization of LLMs in various applications. By providing practitioners with a systematic approach to prompt categorization, we aim to assist in navigating the intricacies of effective prompt design for conversational pre-trained LLMs and inspire new possibilities in their respective fields.
\end{abstract}


\keywords{large language models \and LLMs \and prompt engineering \and prompt design \and prompting techniques \and ChatGPT}


\section{Introduction}




Large Language Models (LLMs)~\citep{radford2019language} have experienced a recent surge in popularity that is fundamentally reshaping the landscape of machine learning~\citep{zhao2023survey}. While early models~\citep{zhou2023comprehensive, bommasani2021opportunities} were characterized by their limited scope and functional capabilities, primarily tailored for specific tasks like text classification and sentiment analysis, the landscape of natural langauge processing (NLP) underwent a transformative shift with the introduction of neural networks and deep learning techniques, enabling models to learn from vast datasets and exhibit adaptability across a broader spectrum of tasks~\citep{petroni2019language, brown2020language}. One pivotal moment in this evolutionary journey was the inception of the transformer architecture~\citep{vaswani2017attention}. This architecture's scalability and remarkable capacity to handle long-range dependencies set the stage for the development of more intricate and powerful models. Subsequently, the introduction of models like BERT~\citep{devlin2018bert} and GPT~\citep{radford2019language} showcased the transformative potential of pre-training these models on extensive datasets, followed by fine-tuning to excel in specific tasks.

\subsection{Conversational Pre-Trained Large Language Models}

In this rapidly evolving landscape, a particular category of LLMs known as conversational decoder-only transformer variants~\citep{vaswani2017attention} like GPT~\citep{brown2020language, ouyang2022training}, LaMDA~\citep{thoppilan2022lamda}, PaLM~\citep{chowdhery2023palm}, LLaMa~\citep{touvron2023llama, touvron2023llama2}, and Mistral~\citep{jiang2023mistral} have emerged as pivotal players in reshaping the field of natural language processing. These models possess an exceptional ability to comprehend, generate, and interpret human language, leading to profound impacts across diverse domains from tech-focused areas like finance~\citep{wu2023bloomberggpt, liu2023fingpt} and programming~\citep{xu2022systematic, roziere2023code} to human-centric sectors like education~\citep{kasneci2023chatgpt, kung2023performance} and healthcare~\citep{abbasian2023conversational, thirunavukarasu2023large}. 
And, due to their conversational nature and the popularity of chat-based interfaces like ChatGPT~\citep{brown2020language, yang2023chatgpt}, the complex computational abilities of these models are readily accessible to nearly anyone with internet access and a cursory interest in artificial intelligence (AI). However, in order for the model to properly respond to a user's query, the user must first know how to communicate with the model by constructing an appropriate prompt.

\begin{figure}
    \centering
    \includegraphics[width=.9\linewidth]{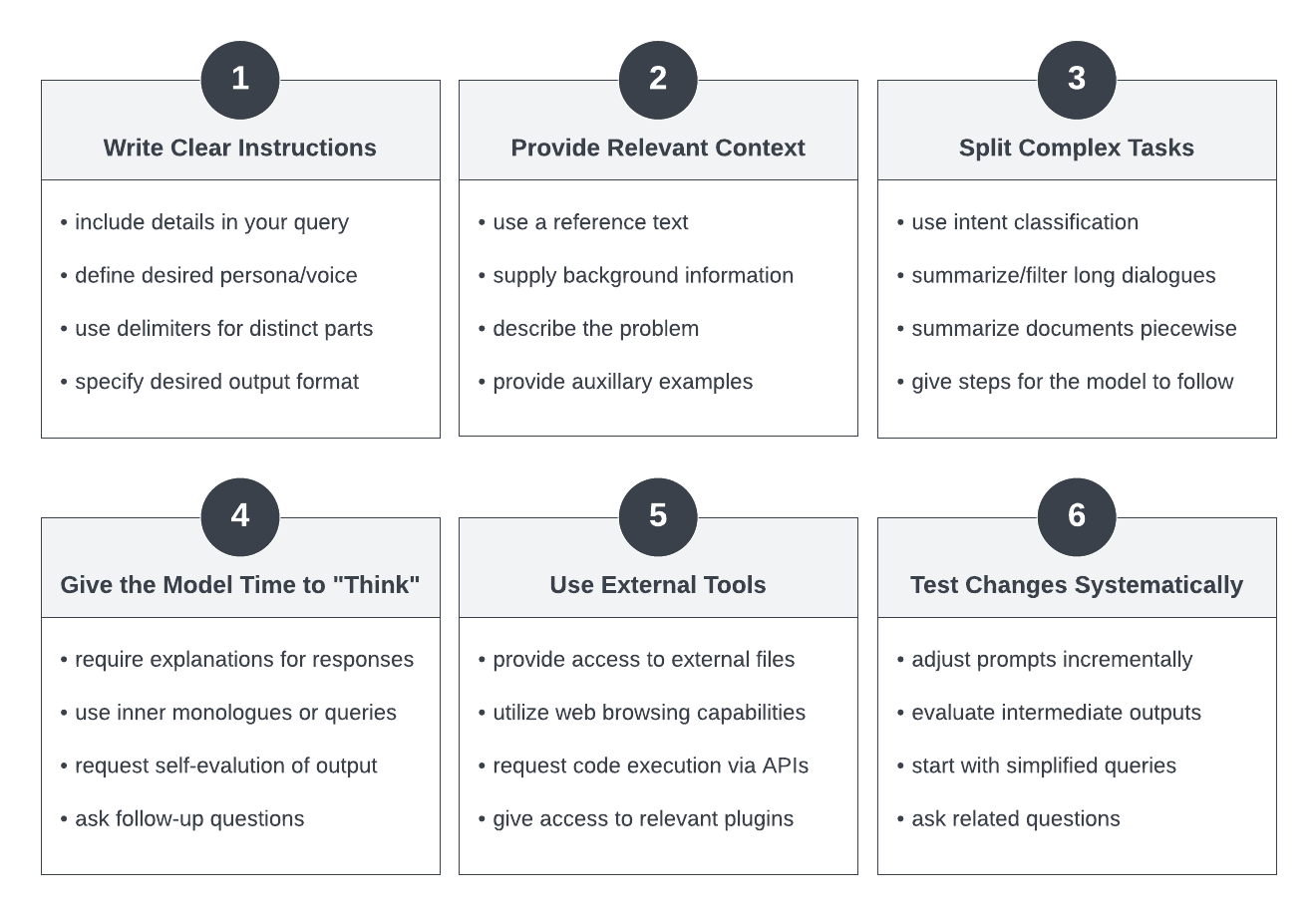}
    \caption{Examples of common best practices for effective prompt design.}
    \label{fig:best_practices}
\end{figure}

\subsection{Prompting for LLMs}
A prompt serves as a translational bridge between human communication and the computational abilities of LLMs. It is comprised of natural language instructions that act as an intermediary language, translating human requests into machine-executable tasks~\citep{liu2023pre}.
Beyond simple instruction, prompts establish a contextual framework for interactions, conveying the significance of information and defining the desired format and content of the model's output.

These instructions can range from straightforward queries like question-answering to more intricate tasks that require the provision of contextual inputs and provide specific requests for the formatting of outputs.

As LLMs continue to evolve, prompts have expanded in scope, enabling novel interaction paradigms that push the boundaries of what these models can accomplish.
With the advent of more sophisticated LLMs like ChatGPT~\citep{bang2023multitask}, the complexity of tasks that prompts are intended to address has also increased.
While the natural language communication avenue of prompts gives the illusion of widespread accessibility, non-AI-experts and individuals with primarily opportunistic approaches to prompting have been shown to be largely inefficient at using LLMs to solve their given tasks~\citep{zamfirescu2023johnny}.
As a result, this rapidly evolving landscape has shifted the focus towards optimizing the strategic use of prompts and created a need for dedicated prompting experts.

\subsection{Prompt Engineering}
Prompt engineering is the strategic process of designing inputs (prompts) that guide LLMs to generate specific, relevant outputs. It's akin to providing the right questions or commands to ensure that the desired answers are obtained. Prompt engineers leverage the predictive capabilities and extensive training data of AI models and combine them with the nuances of human communication to strategically influence model behavior without the need for task-specific training~\citep{brown2020language, petroni2019language, radford2019language, schick2021its}. Effective prompt engineering tactically employs the various best practices of prompt design (see Figure~\ref{fig:best_practices}, derived from OpenAI\footnote{\url{https://platform.openai.com/docs/guides/prompt-engineering}}) as needed to maximize the performance of LLMs, making them more accurate and relevant for specific tasks. It involves not just the creation of prompts but also their testing and refinement based on feedback and desired outcomes.


Furthermore, as tasks asked of LLMs become more complex, effective prompting often necessitates the combination of multiple prompts in new and intricate ways. And since the natural language programming of these models renders them extremely sensitive to changes, even minor modifications in the wording, structure, or context of a prompt can lead to significantly different outcomes from LLMs~\citep{sclar2023quantifying, dong2023revisit}.
Prompt engineers must have a wide knowledge base, not only encompassing how LLMs work but also how each particular prompt pattern fits into the overall wider schema of prompting techniques, and how they each might interact with, improve upon, and affect each other when combined. This requires a nuanced understanding of the strengths and weaknesses associated with each pattern an engineer might consider using, as the quality and structure of prompts directly influence how LLMs process information and produce outputs.
While well-crafted prompts can guide these models to generate more accurate, relevant, and contextually appropriate responses, poorly designed prompts can result in ambiguous, irrelevant, or even incorrect outputs~\citep{brown2020language, ouyang2022training}. The effectiveness of LLMs in practical applications is thus heavily dependent on the quality of prompt engineering, and the expertise of the prompt engineer.

Fortunately, the research literature provides valuable resources for appropriate prompt design by presenting a variety of well-documented prompt patterns suited to a range of use cases. Nonetheless, the abundance of prompting techniques, their diverse presentations, and scattered sources often pose challenges in recalling and navigating through this wealth of information. Therefore, there is a need for a structured categorization of prompting techniques. Such a categorization would not only assist practitioners in making appropriate prompt selections, but also underscore the paramount importance of proper prompt design in achieving successful LLM utilization and ensuring that LLMs produce desired outputs efficiently and effectively in an ever-expanding array of applications.

\begin{figure}
   \centering
   \includegraphics[width=.6\linewidth]{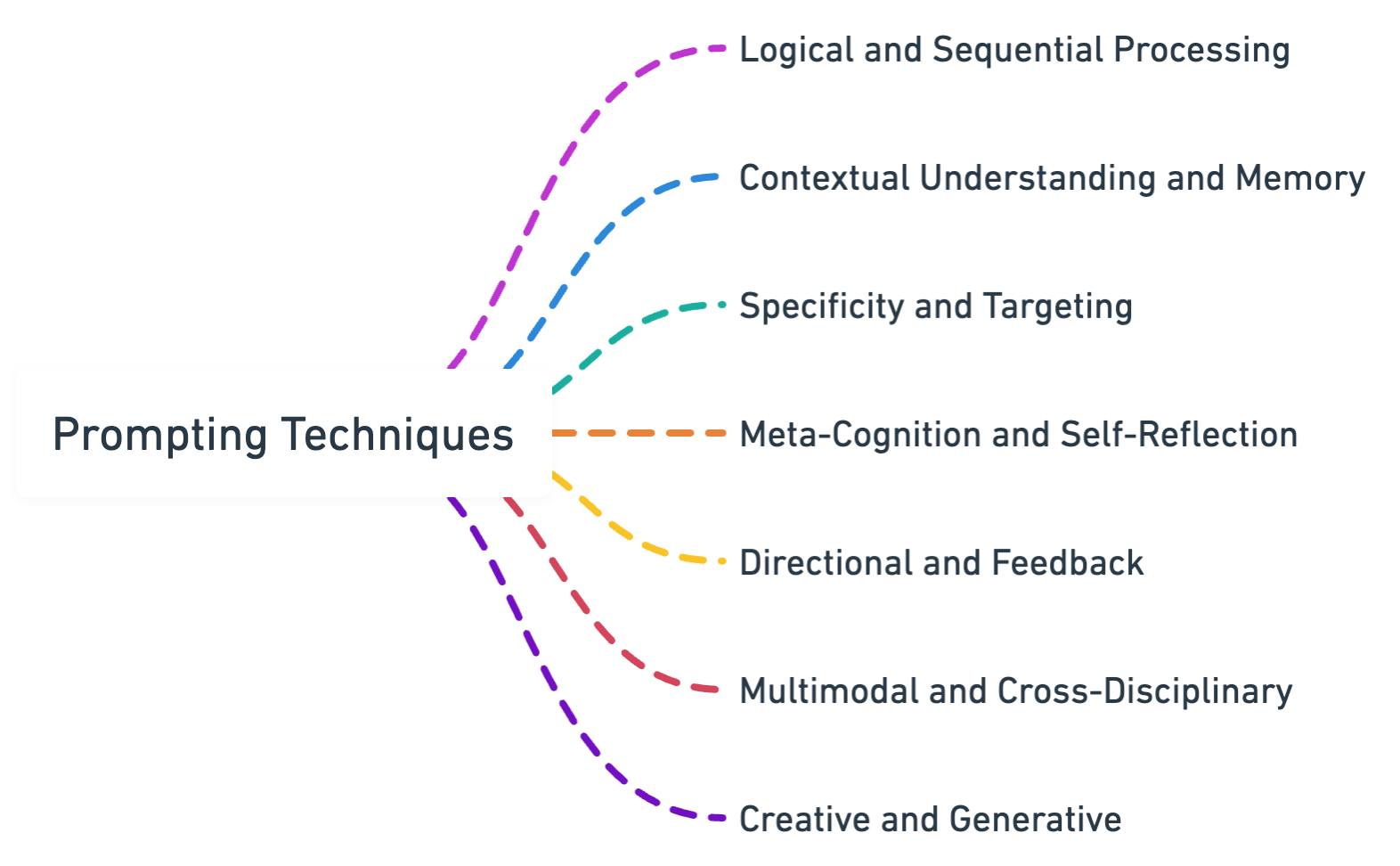}
   \caption{Overview of prompting technique categorization.}
   \label{fig:prompting_categorization}
\end{figure}

\subsection{Framework for Categorization}




In this survey, we present a comprehensive framework for categorizing the diverse range of prompting techniques that have emerged in recent academic literature for conversational Pre-trained Large Language Models (see Figure~\ref{fig:prompting_categorization}).
Our categorization approach is built upon an interdisciplinary foundation, recognizing that the utilization of LLMs spans a varied array of domains and disciplines. We have classified these various techniques and approaches into seven distinct categories, each representing high-level conceptualizations of their approaches and intended use. By organizing these methods into cohesive groups, we offer practitioners a clear roadmap for selecting the most suitable prompting strategies for their specific applications. Whether practitioners are looking to generate creative content, answer questions, or engage in natural language conversations, our framework allows them to easily identify their most relevant categories and subsequently explore techniques that align with their particular goals.

To further enhance the usability of this survey, we have referenced relevant research for each listed approach where applicable to allow for further exploration into the methods of greatest interest. Additionally, we provide real-world examples for all prompting approaches to illustrate how these techniques can be effectively applied in practice. By doing so, we aim to empower practitioners with not only the knowledge of categorization but also the practical insights necessary to implement these techniques successfully.

\begin{figure}
    \centering
    \includegraphics[width=\linewidth]{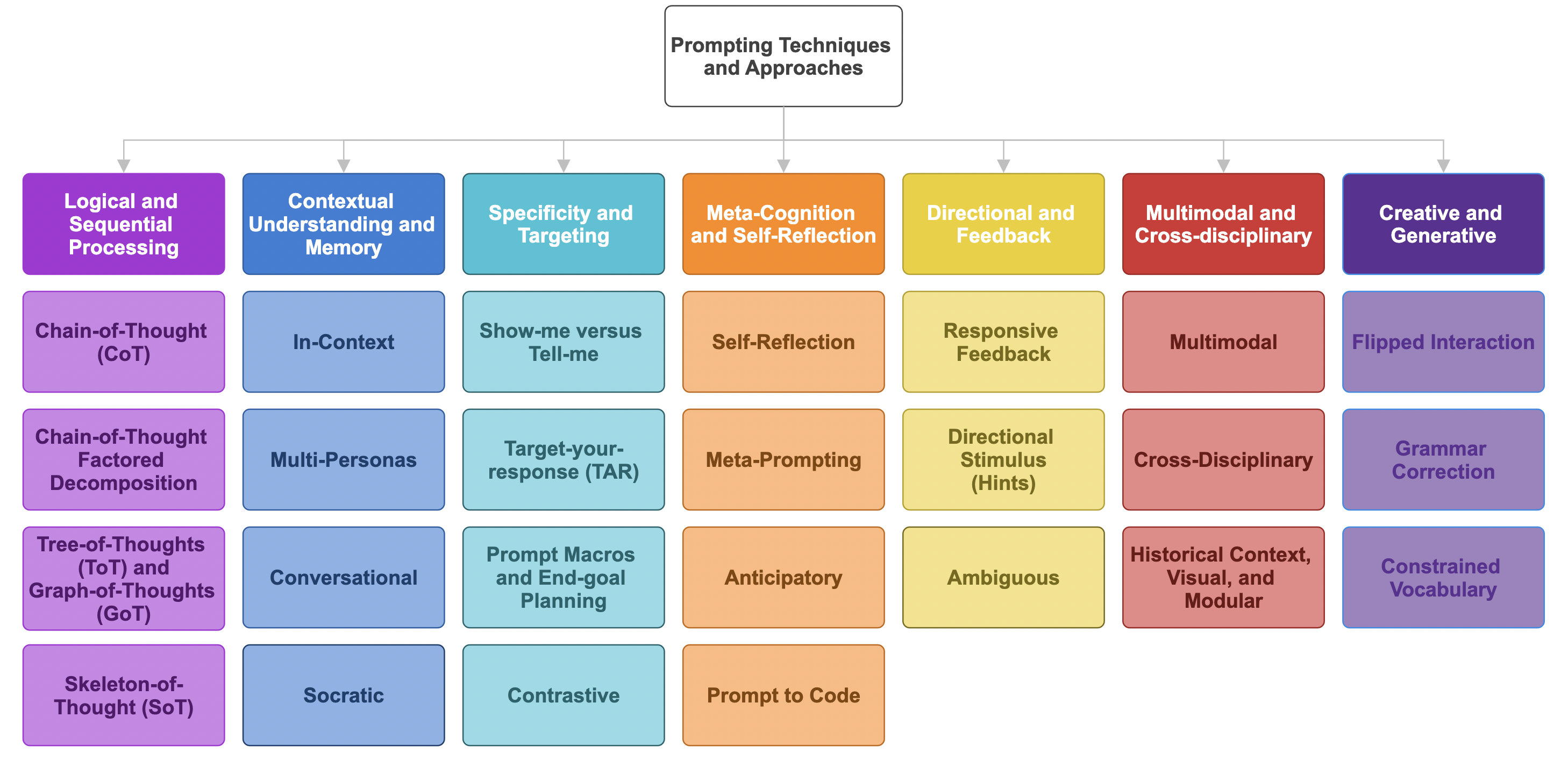}
    \caption{Categorization of prompting techniques and approaches.}
    \label{fig:prompting_techniques}
\end{figure}

The categorization of prompting techniques found in Figure~\ref{fig:prompting_techniques} is as follows:
\begin{enumerate}
    \item \textbf{Logical and Sequential Processing:} Techniques to break down complex reasoning tasks into manageable steps, improving problem-solving and enabling more scientific and human-like reasoning by LLMs.
    \item \textbf{Contextual Understanding and Memory:} Techniques to recall and reference past interactions to deliver seamless conversational experiences and ensure coherence and relevance in extended dialogues.
    \item \textbf{Specificity and Targeting:} Techniques to enhance the precision of LLMs and encourage goal-oriented responses by distinguishing between information types and aligning with specific targets and objectives.
    \item \textbf{Meta-Cognition and Self-Reflection:} Techniques to improve the assistive capabilities of LLMs through introspective prompting, anticipating needs, and generating code.
    \item \textbf{Directional and Feedback:} Techniques to direct the model towards specific tasks or provide feedback that helps the model improve its responses over time.
    \item \textbf{Multimodal and Cross-Disciplinary:} Techniques to integrate diverse input modes and knowledge domains to boost versatility and expand LLM applications across contextually complex scenarios.
    \item \textbf{Creative and Generative}: Techniques to push LLMs to produce creative, engaging, and exploratory content for tasks like storytelling, education, and creative writing.
\end{enumerate}

\section{Categorization of Prompting Techniques and Approaches}

In this section, we present a structured framework for categorizing the diverse array of prompting techniques employed within the domain of Large Language Models. Drawing upon insights from interdisciplinary research, we have systematically organized these techniques into seven distinct categories to offer a comprehensive overview of the current landscape of prompt design in academic literature. Recognizing the broad utility of LLMs across various disciplines, this framework provides a structured lens through which practitioners and researchers can navigate the intricacies of prompt design. Its primary purpose is to facilitate a thorough comprehension of available approaches while also serving as a practical guide for selecting and implementing these techniques in pursuit of specific objectives. By incorporating references to pertinent research and real-world examples, our categorization aims to empower individuals to not only grasp the nuances of effective prompt engineering but also to apply these techniques effectively in practical contexts, advancing the utility of LLMs in natural language processing tasks.

Effective prompt engineering is of critical importance when harnessing the potential of LLMs. The design of prompts significantly influences the behavior of these models, necessitating a structured and intentional approach for prompt selection. This framework serves as a valuable resource, aiding both researchers and practitioners in making informed choices when crafting prompts for a wide range of applications. It encourages a thoughtful consideration of task-specific requirements, enabling users to tailor prompts to achieve optimal results with LLMs while promoting transparency and reproducibility in research and application endeavors involving conversational pre-trained LLMs.



\subsection{Logical and Sequential Processing}

Logical and sequential processing techniques allow an LLM to tackle complex, multi-step reasoning tasks by breaking them down into manageable steps. Methods that fall under this category are based on the idea of splitting the task into smaller components, and have a scientific reasoning and mathematical justification of behind their application. These methodologies not only enhance the models' problem-solving capabilities but also provide a framework for more intuitive and human-like reasoning, opening new avenues for LLM applications across various domains. For a more technical overview of logical and sequential processing techniques, we refer the reader to the survey~\citep{yu2023towards} and the references therein. 

Techniques introduced in this section include: Chain-of-Thought (CoT) Prompting, Chain-of-Thought Factored Decomposition Prompting, Tree-of-Thoughts (ToT) and Graph-of-Thoughts (GoT) Prompting, and Skeleton-of-Thought (SoT) Prompting. 




\subsubsection*{Chain-of-Thought (CoT) Prompting}

Chain-of-Thought (CoT) prompting causes the model to ``think'' in a sequential manner by processing information step-by-step~\citep{wei2022chain, kojima2022large, lyu2023faithful, wang2023knowledge, wang2023boosting}. Rather than jumping straight to a conclusion, this technique prompts the model to take a moment to dissect complex queries into explainable intermediary steps. By having the model ``show its work'' and logically reason through each step of problem solving, the overall accuracy of its solutions is generally seen to increase.

\textit{Practical Example}: You are looking to have the model solve a math problem. Instead of simply asking the model ``What's the result of integrating $x^2$ from $0$ to $1$?'' you could prompt sequential thinking within the same prompt by also adding: ``Start by defining the function. Next, set the integration limits. Now, walk me through the integration process step by step.'' This way, you can follow the model's logical progression, and the model is given time to process each step appropriately.

\subsubsection*{Chain-of-Thought Factored Decomposition Prompting}
Chain-of-Thought Factored Decomposition Prompting combines the sequential reasoning inherent to chain-of-thought prompting with \textit{factored decomposition}: the breaking down of complex tasks or ideas into smaller, more manageable parts~\citep{wang2022self, fu2022complexity, xue2023rcot}. This technique refines the model's analytical approach by guiding it to think sequentially while also breaking down and addressing each subcomponent of the given task.



\textit{Practical Example}: You would like the model to explain the process of photosynthesis. Instead of issuing a direct query (e.g. "How does photosynthesis work?"), you might structure the prompt to first ask the LLM to define all the main processes involved in photosynthesis before detailing the steps sequentially. This would help ensure that each aspect of photosynthesis is thoroughly decomposed and explained in an accessible way.

\subsubsection*{Tree-of-Thoughts (ToT) and Graph-of-Thoughts (GoT) Prompting}
Tree-of-Thoughts (ToT) and Graph-of-Thoughts (GoT) prompting techniques build on CoT prompting to expand beyond linear reasoning and explore more diverse and creative pathways of thinking.


ToT~\citep{yao2023tree} builds upon CoT by forming a decision tree by which to guide the model through various logical paths and enhance its decision-making processes. For scenarios like brainstorming sessions, ToT enables the model to simulate the process of mind mapping to explore multiple branches of a core idea.

Similarly, GoT models LLM prompts as graphs, facilitating intricate thought aggregation and manipulation~\citep{besta2023graph}. This approach allows for enhanced modular and dynamic reasoning.

\textit{Practical Example}: You would like the model to help you brainstorm business strategies. In this instance, one might start with a core idea (e.g. improving customer service), and then ask the AI to generate multiple branching strategies or ``What if?'' scenarios through which to explore the various possibilities.

\subsubsection*{Skeleton-of-Thought (SoT) Prompting}
Skeleton-of-Thought (SoT) prompting parallels human problem-solving by providing a segmented yet comprehensive framework for responses~\citep{ning2023skeleton}. By giving the model a structured, high-level ``skeleton'' of the desired response, it is now able to ensure proper filling of the provided template without generating content too far outside the bounds of the desired output.



\textit{Practical Example}: You would like the model to draft a business email. As you already know how you would like the email to be structured, you could provide the following template: ``[Greeting], [Introduce topic], [Main content], [Closing remarks], [Signature].'' The model is thereby prompted to fill in each of the requested sections without foregoing important parts of the template or adding more than was initially desired.

\subsection{Contextual Understanding and Memory}
LLMs have evolved to exhibit advanced levels of contextual understanding and memory, enabling them to maintain relevance and coherence throughout extended dialogues. This capability is achieved through various innovative techniques that allow these models to recall and reference previous interactions, thus offering a seamless conversational experience.

Techniques introduced in this section include: In-Context Prompting, Multi-Personas Prompting, Conversational Prompting, and Socratic Prompting.

\subsubsection*{In-Context Prompting}
In-context prompting emulates a form of ``memory'' for the model by maintaining context over multiple interactions. This technique integrates historical context into current responses in order to leverage the significance of attention mechanisms in enhancing the coherence of sequences in LLMs~\citep{rubin2021learning, dong2022survey, min2021metaicl, ye2022complementary, wang2022iteratively}. By providing information to an LLM within its current context window, prompt engineers are able to build upon previously introduced information to construct more complex ideas and scenarios and generate more coherent and context-aware responses.




\textit{Practical Example}: You are conversing with a virtual assistant LLM about planning a trip. You initially mention you are traveling to Paris, and later ask about ``recommended restaurants.'' Due to the conversation happening within the model's context window, the model will remember your past interactions, and is likely to suggest eateries in Paris.

\subsubsection*{Multi-Personas Prompting}
Multi-personas prompting utilizes an LLM's ability to exhibit consistent character or persona traits to make a model ``wear different hats'' during an interaction~\citep{gu2023effectiveness}. This technique may enhance user experience by offering a stable conversational partner who is simultaneously adaptable to various voices,  perspectives, and expertise.





\textit{Practical Example}: You would like to see how differently Shakespeare and a scientist might describe a sunset. You might first ask: ``How would Shakespeare describe a sunset?'' After receiving a response, you might then follow up with: ``Now how would a scientist describe a sunset?'' Even though the subject (sunset) remains constant, the model is able to switch between the poetic persona of Shakespeare and the analytical persona of a scientist, thus providing you with two different perceptions of the same topic.

\subsubsection*{Conversational Prompting}
Conversational prompting involves crafting prompts that mimic natural human conversation~\citep{wei2023leveraging, liu2023icsberts}. This not only improves the fluidity of the model's responses but also increases their relevance. This is evident when a model engages in interactive dialogue with the user, such as inquiring about additional details or suggesting follow-up questions, thereby enriching the conversational experience. Encouraging back-and-forth responses can lead to richer, more nuanced interchanges with a model, and marks a transformative shift from seeing AI interactions as singular queries to viewing them as ongoing dialogues.



\textit{Practical Example}: You're packing for an upcoming trip. Instead of asking the singular question, ``What is the weather like in Athens?'', you might continue to follow up with, ``What should I pack for a day of sightseeing?'' or ``Any cultural norms I should be aware of?'' in order to foster a more thorough, insightful, and freeform experience.

\subsubsection*{Socratic Prompting}
Socratic prompting emulates the Socratic method of dialogue by asking a series of questions to lead the model (or the user) to a conclusion or realization~\citep{chang2023prompting}. This technique allows the user to explore the depth of knowledge an LLM has around a certain topic by probing into particular areas of interest.



\textit{Practical Example}: You are curious about the concept of justice. You may start prompting the model with a general question (e.g. ``What is justice?''), and then following up with more nuanced questions like ``Is it always aligned with legality?'' based on the model's answers.

\subsection{Specificity and Targeting}
The development of specificity and targeting techniques for LLMs significantly enhances their ability to produce precise, goal-oriented responses. By enabling these models to distinguish between different types of information delivery, focus on specific response targets, and align with overarching objectives, these approaches vastly improve the utility and applicability of LLMs in various domains requiring detailed and targeted information processing.


Techniques introduced in this section include: Show-me versus Tell-me Prompting, Target-your-response (TAR) Prompting, Prompt Macros and End-goal Planning, and Contrastive Prompting.

\subsubsection*{Show-me versus Tell-me Prompting}
Show-me versus tell-me prompting involves instructing LLMs to either demonstrate (``show'') or describe (``tell'') indicated concepts~\citep{perez2021true}. The success of this approach is predicated on the user's ability to discern what type of information would be most valuable to them, or which output would best suit the task at hand. For example, a ``show-me'' prompt might request a visual demonstration, while a ``tell-me'' prompt could seek a descriptive explanation of the same process. Utilizing both methods simultaneously or back-to-back can be used demonstrate the model's ability to efficiently understand and adapt to nuanced differences in requests for forms of information delivery.



\textit{Practical Example}: You would like to know how photosynthesis works. You might ask the model to either ``Show me how photosynthesis works with a diagram,'' or ``Tell me about the process of photosynthesis.'' While the former request might yield a graphical representation, the latter would likely give a textual explanation, and your ideal formulation of output paired with predictions of which output formats the model might deliver should both inform the initial request.

\subsubsection*{Target-your-response (TAR) Prompting}
Target-your-response (TAR) prompting directs the model to focus its responses toward specific targets or objectives in order to enhance the relevance and brevity of its output~\citep{zhao2021calibrate}. This method is about fine-tuning the specificity of the desired answer. It emphasizes the importance of clearly indicating the format or style of response you would like to recieve.



\textit{Practical Example}: You would like to know more about coffee. Instead of simply asking ``Tell me about coffee,'' you might instead say: ``Provide a 3-sentence summary about the origins of coffee.'' The latter prompt more carefully guides the model to a specific query and provides enhanced contextual bounds for its response.

\subsubsection*{Prompt Macros and End-goal Planning}
Prompt macros and end-goal planning utilize pre-defined prompt templates to establish overarching goals for the model to reach with its responses. They ensure the LLM serves the broader goal with each interaction by combining potentially numerous ``micro'' prompts and queries into a single larger, ``macro'' prompt. In designing a macro prompt, you must ensure that the prompt is broad enough to encompass the breadth of all desired micro queries, but also narrow enough that the model understands all relevant specifics of the implied requests.



\textit{Practical Example}: You would like the model to help you plan a trip within a single interaction. Rather than asking individual questions one-by-one, you might choose to use a macro prompt like ``Plan a 5-day trip to Colorado in January,'' where the model is likely to interpret this request with the appropriate level of nuance to provide itinerary, lodging, and activity suggestions in sequence.

\subsubsection*{Contrastive Prompting}
Contrastive prompting involves asking the model to compare or contrast two concepts, objects, or ideas. By framing a prompt in this manner, the model is tasked with identifying differences or similarities between the provided subjects, thereby allowing for a deeper understanding of their characteristics and relationships. This approach leverages the model's ability to discern subtleties and enables it to provide more insightful and contextually relevant responses.


\textit{Practical Example}: You would like to improve your understanding of the differences between two programming languages, Python and Java. You might prompt the model to ``Compare and contrast Python and Java programming languages,'' in order to gain insights into their syntax, performance, and use cases. This approach can help you make informed decisions about which language to choose for your specific project in order to streamline your development process and optimize your software's performance.

\subsection{Meta-Cognition and Self-Reflection}

The Meta-Cognition~\citep{wang2023metacognitive}, Self-Reflection~\citep{shinn2023reflexion}, and Self-Explanation~\citep{gao2023self} domain delves into the self-analytical capabilities of LLMs.
By empowering these models to engage in self-guided learning, anticipate user needs, and generate programming code, these methods not only broaden the scope of LLM applications but also enhance their interactive and assistive capabilities, making them indispensable tools in various technology-driven fields.


Techniques introduced in this section include: Self-reflection Prompting, Meta-Prompting, Anticipatory Prompting, and Prompt to Code.


\subsubsection*{Self-reflection Prompting}

Self-reflection prompting allows the model to evaluate its own responses and critically process its answers~\citep{shinn2023reflexion, madaan2023self, xie2023decomposition}. This is particularly helpful when dealing with intricate tasks that were not initially prompted in a manner conducive to the model's ability to work through each step methodically. By prompting the LLM to self-reflect on previous outputs, the model is able to review and update content so that user receives more deliberate and thoughtful responses.






\textit{Practical Example}: A model has given a questionable answer on an ethical matter. To have the model reassess and add more color to its initial answer, you might follow up with, ``Are you sure about that?'' This would prompt the model to self-reflect on its previous response and potentially lead to both a closer inspection of the initial question and either a newly updated answer or a detailed reasoning for the first.

\subsubsection*{Meta-Prompting}
Meta-prompting involves guiding LLMs to reflect on their own processes and methodologies for responding to prompts~\citep{wang2023metacognitive, gao2023self}. This approach not only enhances the model's understanding of its capabilities within the current context window, but also improves its interaction quality by encouraging a level of self-awareness and introspection into the process of interacting with LLMs.



\textit{Practical Example}: You would like to know how to write better prompts, so you ask the model to ``Analyze the effectiveness of the last 5 prompts given to you and suggest improvements.'' Now that the process of prompting has been introduced into the current context, the model may provide insights into improving the design of your prompts to maximize beneficial outputs from LLMs.

\subsubsection*{Anticipatory Prompting}
Anticipatory prompting enables AI models to foresee and address future queries or needs based on the current context. This method involves crafting prompts that encourage the model to provide both direct answers as well as related insights, thereby anticipating possible follow-up questions or concerns.



\textit{Practical Example}: When asked, ``How do I plant tomatoes?'' the model might extrapolate from the context that you are new to growing tomatoes, and, as a result, also provide tips on dealing with common pests in anticipation of a potential follow-up concern.

\subsubsection*{Prompt to Code}
Prompt to code focuses on instructing the AI to generate functional programming code based on specific prompts~\citep{chen2023teaching}. It capitalizes on the programming information contained within the model's training data in order to understand and produce code snippets according to the language, formatting, and other requirements delineated by the user.



\textit{Practical Example}: You would like to have a model generate code for a development project. You might ask the LLM to ``Generate a Python function to calculate the Fibonacci series for a given number.'' With this direction, the model would then be able to produce the relevant code according to your specifications.

\subsection{Directional and Feedback Techniques}

Directional and feedback techniques guide an LLM towards specific tasks or refine its responses based on user feedback.
By employing responsive feedback, users can actively participate in the model's learning process, ensuring outputs are more aligned with their expectations and needs. Simultaneously, directional stimulus prompting fosters a balance between guided output and AI autonomy, encouraging creative and tailored responses. These methodologies not only refine the interaction between users and models, but also significantly enhance the LLM's adaptability and precision in various applications.

This section introduces the following techniques: Responsive Feedback Prompting, Directional Stimulus Prompting, and Ambiguous Prompting.

\subsubsection*{Responsive Feedback Prompting}
Responsive feedback prompting incorporates feedback directly into the prompting process to improve the quality and relevance of LLM responses~\citep{zheng2023progressive, paul2023refiner}. In this approach, feedback is given immediately after the initial model output in order to guide subsequent responses.


\textit{Practical Example}: You are using a model to brainstorm ideas for a logo. The model has produced a mostly satisfactory result, but you would like a few changes. Prompting the model with feedback that states ``I like the color scheme, but can the design be more minimalistic?'' gives relevant input that allows the model to modify its next output to be more in line with the user's preferences.


\subsubsection*{Directional Stimulus Prompting}
Directional stimulus prompting involves using hints or subtle cues to steer the LLM in a desired direction without explicitly dictating the output~\citep{shum2023automatic}. This technique is particularly useful when desiring unexpectedness and enhanced creativity in the model's response. By only hinting at what the user desires in the output, the model is left to fill in the blanks and make a best guess.


\textit{Practical Example}: You are prompting a model to generate a story. Rather than describing
the plot line in detail and receiving a story entirely in-line with your requests, it might be more interesting to include a hint like ``Add an element of surprise when the hero meets the princess.'' This allows the user to indicate that they would like to have something unexpected happen without requiring them to specify exactly what that something is, and nudges the LLM to incorporate an unspecified twist while leaving the user in suspense of the specifics.


\subsubsection*{Ambiguous Prompting}
Ambiguous prompting is the intentional use of vague prompts to stimulate creativity or broad spectrums of responses from the model. This method is similar to Directional Stimulus Prompting, but intentionally designed to be even more open-ended to encourage a substantial level of creativity and unguided generation. Ambiguous prompts are best used when the user is either unsure of what they're looking for in a response and/or would like to see unbiased and relatively unprompted ideas.



\textit{Practical Example}: You are prompting a model to generate a story. Instead of requesting a specific narrative through a detailed prompt like ``Write a story about a knight saving a princess from a dragon,'' it might be more interesting to instead request that the model simply ``Write a story about bravery.'' By keeping the prompt vague, you reduce the amount of context given to the LLM and instead encourage the output of a narrative uninfluenced by specific user preferences.


\subsection{Multimodal and Cross-Disciplinary Techniques}
The Multimodal and Cross-Disciplinary category encompasses techniques that integrate various modes of input and diverse knowledge domains to enhance the versatility and applicability of LLMs. By combining multiple input types and blending knowledge from various disciplines, these techniques not only cultivate the depth and breadth of LLM responses but also open up new possibilities for applications in diverse and complex scenarios, ranging from artistic endeavors to scientific research and historical analysis.

Techniques introduced in this section include: Multimodal Prompting, Cross-disciplinary Prompting, and Historical Context, Visual, and Modular Prompting.


\subsubsection*{Multimodal Prompting}
Multimodal prompting refers to the use of diverse input types in prompting LLMs~\citep{bang2023multitask, lee2023multimodal}. Now able to process more than just text, some models can receive prompts containing various combinations of words, images, audio files, and even videos to help provide context and guidance for the LLM's output.



\textit{Practical Example}: You would like a model to write a poem about a particular photo of a sunset. It is important to you that the model incorporate details specific to this image. With a model capable of receiving multimodal inputs, you might upload your photo of a sunset and simultaneously ask the model to ``Describe this scene by writing a poem.'' By processing the photo with computer vision capabilities and the user request through the LLM, the model will have an understanding of both the visual and textual inputs in order to complete the request.

\subsubsection*{Cross-disciplinary Prompting}
Cross-disciplinary prompting involves blending knowledge from multiple separate disciplines to prompt unique solutions to interdisciplinary problems~\citep{cui2023chatlaw, harrison2023zero, huang2022inner}. By equipping LLMs with richly tailored tools and knowledge, they become more adept at handling complex queries and tasks that necessitate the inclusion of insights from various domains. This not only enhances their accuracy and relevance in these targeted areas, but also opens up new possibilities for their application in various professional and academic fields more generally.





\textit{Practical Example}: You are an avid fiction reader taking a class in physics. You struggle to understand some of the more complicated scientific topics, as they're conceptually very different from the literature you're most interested in. You might prompt a model to ``Explain quantum physics principles using analogies from classical literature.'' By asking the model to explain one discipline through the lens of another, you expect the model to understand both domains well enough to provide an nuanced yet accurate interpretation of disparate concepts in order to connect the dots in your understanding.

\subsubsection*{Historical Context, Visual, and Modular Prompting}
Historical context, visual, and modular prompting techniques focus on embedding historical context, visual elements, and modular constructs into prompts~\citep{liu2023pre}. A historical context prompt might ask the model to answer in accordance with a particular historical setting or reference, while visual prompting might involve using images to guide the model's responses, and modular prompting leverages structured, component-based prompts to address complex queries.



\textit{Practical Example}: You're curious about how your great-grandfather might view modern technology, and could prompt the LLM to ``Describe the Internet as someone from the 1920s would understand it.'' By specifying the historical setting (in this case, a particular decade) for the output, the model is able to better contextualize its response.

\subsection{Creative and Generative Techniques}

Creative and generative techniques enable LLMs to generate creative content and elicit diverse, innovative, and thought-provoking responses. By employing a range of prompting strategies, from encouraging ambiguity to imposing lexical constraints, these techniques not only expand the capabilities of LLMs in creative tasks but also open up new avenues for their application in diverse domains like storytelling, educational content generation, and creative writing.



Techniques introduced in this section include: Flipped Interaction Prompting, Grammar Correction, and Constrained Vocabulary Prompting.





\subsubsection*{Flipped Interaction Prompting}
Flipped interaction prompting reverses the conventional model-user dynamic. Instead of the user leading and directing the conversation with the model, the model might instead pose questions or seek clarifications from the user. By switching up the expected roles of the interaction, the model is able to guide the user to create their own outputs through targeted queries.
This back-and-forth is helpful for structured, conversational brainstorming sessions and allows the user to use the LLM to prompt their own development of responses so that they maintain full creative control over the outputs of the interactions.



\textit{Practical Example}: You are drafting a business plan. Rather than providing details to the model and having it generate suggestions, the user might prompt the model to ask them questions to guide their own creation of the business plan by telling the model to ``Ask me questions one-by-one to guide me in creating a comprehensive business plan.'' In response, the model might begin asking ``What's your business's main objective?'' or ``Who is your target audience?'' in order to prompt the the user to consider various aspects of business plan development.


\subsubsection*{Grammar Correction}
Grammar correction prompts are specific instructions given to a model to identify and rectify grammatical errors or inconsistencies in the text provided by a user. As LLMs are trained on vast corpora of textual data, these prompts are able to leverage the model's well-versed language understanding capabilities to serve as a conversational grammar checker, offering suggestions and improvements to the user's input according to the requested output requirements. By asking the LLM to make improvements to verbiage and writing style, users can enhance the quality and clarity of their written communication and easily adapt their writing to various use cases and contexts.




\textit{Practical Example}: You are writing a formal report and want to ensure that your grammar and tone is appropriate for the given context. You might ask the model to ``Review the following report for proper grammar and ensure that the language is clear, but professional.'' By requesting both tonal and grammatical improvements, the model is given appropriate contextual bounds by which it may analyze and adjust the provided text.

\subsubsection*{Constrained Vocabulary Prompting}
Constrained vocabulary prompting involves restricting the model's response to a specific set of words or a defined vocabulary. This technique places tangible limitations on the desired output of the model, ensuring that the generated content adheres to a predefined lexicon. By constraining the vocabulary, we gain greater control over the language generation process, making it particularly useful in scenarios where precision, adherence to specific terminology, or the avoidance of sensitive or inappropriate language is of utmost importance.



\textit{Practical Example}: You are developing a chatbot for a customer service application in the healthcare industry. To ensure the bot provides accurate and consistent responses, you might instruct the model to ``Answer customer queries using only medical terminology and avoid colloquial language.'' This approach encourages the model to maintain a professional and medically accurate tone, reducing the risk of misunderstandings or misinformation when interacting with users seeking healthcare-related information or assistance.


\section{Discussion}

In this work, we connect our findings to existing research to highlight important aspects of practical prompt engineering. Despite recognizing the limitations of the field and the myriad active developments it contains, we believe our categorization framework can be useful to practitioners by providing a resource for prompt selection and encouraging discussion around potential use cases for each prompting method presented. By examining the related research and considering potential ethical challenges and limitations, this section consolidates our survey's overarching narrative in guiding the reader toward a comprehensive understanding of the nuanced aspects of LLM prompting.

\subsection{Related Work}
A number of surveys have been previously conducted to categorize prompting techniques into various frameworks. Software patterns~\citep{schmidt2013pattern} in particular have been
used as a framework for explaining and detailing prompting techniques,  providing examples, key ideas, context, and motivation for use~\citep{white2023prompt}. As these software patterns provide reusable solutions to recurring problems, prompting techniques are often well-suited to such a formulaic breakdown, especially for prompts that are notably structured in nature or otherwise designed for more technical use cases.

Similarly, other catalogs of prompting methods have provided a technical look at prompting and prompt-based learning techniques~\citep{liu2023pre, kanti2023teler}, with some focusing specifically on the context of engineering for software-reliant systems~\citep{schmidt2023towards}. These approaches reflect the broader trend of catering to more technically inclined practitioners in the current landscape of prompting technique surveys.

Due to the widespread attention towards LLMs brought about by commonly accessible chatbot interfaces like ChatGPT, the importance of good prompt design for LLMs has seen substantial interest in recent years~\citep{van2023chatgpt, bang2023multitask, yang2023harnessing}. Several works have contributed to the evolving landscape of prompt techniques by investigating innovative ways of approaching prompt design as a whole. For example, some studies have looked into the aggregation of multiple prompts~\citep{arora2022ask}, while other work suggests that zero-shot prompting methods significantly outperform few-shot approaches~\citep{reynolds2021prompt}, and others propose automated prompt generation as a parameter-free alternative to traditional finetuning methods~\citep{shin2020autoprompt}.

Additionally, as further research delves into unraveling the full potential of pre-trained LLMs, it has prompted an examination of our contemporary understanding of prompts and the role of human-led prompt engineering. Investigations into emergent abilities in scaled LLMs~\citep{wei2022emergent}, automated prompt generation as a parameter-free alternative to traditional fine-tuning methods~\citep{shin2020autoprompt}, and the use of Automatic Prompt Engineers~\citep{zhou2022large} collectively represent a reevaluation of traditional prompt paradigms. Furthermore, techniques like chain-of-thought prompting have been regularly studied as means by which to enhance LLM reasoning abilities in multistep problem solving tasks without the necessity of further user inputs~\citep{wei2022chain, wang2022self, kojima2022large, wang2023knowledge}. Together, these studies encompass various strategies that collectively contribute to the potential for future model-led optimizations of LLM outputs for a wide array of applications.



\subsection{Challenges and Ethical Considerations}
Ethical considerations in prompt design are critical to the development and application of large language models~\citep{tokayev2023ethical}. While these models exhibit impressive capabilities, they also raise significant ethical concerns that must be addressed to ensure responsible and safe usage. One central issue revolves around the potential for biased outputs~\citep{bender2021dangers, o2017weapons}, reflecting discussions on algorithmic fairness and ethics. Addressing this challenge requires careful consideration of not only the training data and models but also the design of prompts themselves.

It is essential to acknowledge that LLMs are influenced by the data they are trained on, which can introduce biases. These biases may manifest in the form of stereotypes, prejudices, or unfair representations of certain groups or ideas~\citep{zhuo2023exploring, tokayev2023ethical}. Consequently, prompt designers must be acutely aware of the potential pitfalls and actively work to mitigate bias in both prompts and their generated responses. This involves scrutinizing and refining prompts to avoid reinforcing harmful stereotypes or producing discriminatory content.

Furthermore, ethical prompting extends beyond mere avoidance of bias. It necessitates crafting prompts that align with societal norms and values, fostering fairness and positivity in LLM responses~\citep{shaikh2022second, xu2023llm}. This alignment seeks to ensure that the outputs generated by these models are in line with ethical principles and contribute positively to the well-being of individuals and society as a whole. Designers must carefully consider the consequences of prompts and strive to guide LLMs toward producing responsible and socially acceptable content~\citep{ma2023oops}.

In the realm of prompt design, another ethical dimension involves addressing illicit and invasive prompts~\citep{deng2023multilingual}. Illicit prompts are those that seek information for potentially harmful purposes, such as queries related to illegal activities or harm-inducing instructions. Invasive prompts, on the other hand, risk breaching privacy or confidentiality, which can have serious ethical implications. Prompt designers must exercise vigilance in identifying and preventing such prompts, ensuring that they do not lead to the generation of harmful or inappropriate content~\citep{wolf2023fundamental}.

To bolster ethical prompting, the implementation of trust layers is crucial. These mechanisms serve as safeguards to ensure that AI-generated content is reliable, interpretable, and trustworthy~\citep{zafar2023building, mokander2023auditing}. Trust layers involve cross-checking AI responses against ethical standards, accuracy benchmarks, and organizational objectives. They provide an additional layer of assurance, helping to maintain trust in AI technologies and their responsible and beneficial use across various domains like business and technology.

Ethical considerations in prompt design for LLMs are pivotal to their responsible and safe deployment. By diligently addressing biases, avoiding illicit and invasive prompts, and implementing trust layers, prompt engineers can contribute to the development of AI systems that operate within the boundaries of moral responsibility and societal acceptability. These measures are essential not only for maintaining trust in AI technologies but also for ensuring their sustainable and positive impact on society.




\subsection{Concluding Remarks}
In this survey, we navigate the landscape of prompting techniques and approaches for conversational pre-trained large language models and offer a comprehensive overview accessible to both casual users of chatbot-style models and seasoned prompt engineering professionals. Our goal is to bridge the gap between academia and practical applications, presenting a non-technical perspective on the diverse approaches prevalent in both research and real-world usage. Unlike traditional surveys that may delve deeply into technical nuances, our categorization is tailored to resonate with a broader, more interdisciplinary audience, ensuring that even those without a specialized background in artificial intelligence and machine learning can grasp the fundamental concepts and techniques presented herein.

Our survey represents a valuable contribution to the field of large language models by offering a novel perspective on categorizing prompting techniques. While we have maintained a commitment to accessibility and interdisciplinary appeal, the significance of our categorization extends beyond its user-friendly approach. By providing a structured framework for understanding and employing these techniques, we aim to empower a wider range of users to tap into the transformative potential of large language models in their respective fields. Moreover, our categorization encourages a deeper exploration of the creative applications and innovative solutions that can emerge when diverse perspectives and backgrounds intersect with the capabilities of these models. In essence, this work serves as a catalyst for fostering cross-disciplinary collaborations and driving forward the democratization of language model utilization, ultimately enriching the overall landscape of human-AI interaction.



\setcitestyle{numbers}
\bibliographystyle{abbrvnat}
\bibliography{references}

\end{document}